\documentclass{article}

\usepackage{arxiv}

\usepackage[utf8]{inputenc} 
\usepackage[T1]{fontenc}    
\usepackage{hyperref}       
\usepackage{url}            
\usepackage{booktabs}       
\usepackage{amsfonts}       
\usepackage{nicefrac}       
\usepackage{microtype}      
\usepackage{lipsum}

\usepackage[section]{placeins}
\usepackage[shortlabels]{enumitem}
\usepackage{subcaption}
\usepackage{cite}
\usepackage{amsmath,amssymb,amsfonts}
\usepackage{algorithmic}
\usepackage{graphicx}
\usepackage{textcomp}
\usepackage{xcolor}
\def\BibTeX{{\rm B\kern-.05em{\sc i\kern-.025em b}\kern-.08em
    T\kern-.1667em\lower.7ex\hbox{E}\kern-.125emX}}
    
\usepackage{verbatim}  

\usepackage{hyperref}
\usepackage{csquotes}
\usepackage{todonotes}
\usepackage{float}

\usepackage{ tipa }

\usepackage[utf8]{inputenc}

\title{Towards self-organized control: Using neural cellular automata to robustly control a cart-pole agent}

\author{Alexandre Variengien$^{1,2,3}$, Sidney Pontes-Filho$^{1,4}$, Tom Glover$^{1}$ and Stefano Nichele$^{1,2,*}$
\bigskip \\ 
\small
$^1$Department of Computer Science, Oslo Metropolitan University, Oslo (Norway)\\
$^2$ Department of Holistic Systems, Simula Metropolitan Centre for Digital Engineering, Oslo (Norway)\\
$^3$Department of Computer Science, École Normale Supérieure de Lyon, Lyon (France)\\
$^4$Department of Computer Science, Norwegian University of Science and Technology, Trondheim (Norway)\\
$^*$Corresponding author: stenic@oslomet.no}

\date{\small

}

\begin{document}
\maketitle
\begin{abstract}
   Neural cellular automata (Neural CA) are a recent framework used to model biological phenomena emerging from multicellular organisms. In these systems, artificial neural networks are used as update rules for cellular automata. Neural CA are end-to-end differentiable systems where the parameters of the neural network can be learned to achieve a particular task. In this work, we used neural CA to control a cart-pole agent. The observations of the environment are transmitted in input cells while the values of output cells are used as a readout of the system. We trained the model using deep-Q learning where the states of the output cells were used as the q-value estimates to be optimized.
    We found that the computing abilities of the cellular automata were maintained over several hundreds of thousands of iterations, producing an emergent stable behavior in the environment it controls for thousands of steps. Moreover, the system demonstrated life-like phenomena such as a developmental phase, regeneration after damage, stability despite a noisy environment, and robustness to unseen disruption such as input deletion.
\end{abstract}

\section{Introduction}
 One of the most remarkable feats of life is the developmental process leading to the emergent complexity of the human brain from a single cell. The field of neurodevelopment (i.e., the development of the nervous system) has been investigating this problem for decades. These studies led to the discovery of the intricate mechanisms by which gradients of chemicals and local cell interactions shape the differentiation of pre-neural cells and the organization of their connections \cite{sansom2009gradients}.

Even after the brain is considered fully grown, its developmental process does not stop. New neurons are formed continuously until death, and the shape and the strength of their connections change. Such neural plasticity is influenced by the sensory inputs received by the individual and is considered to be at the root of the emergence of intelligence. In addition to the ability to cope with a changing environment, plasticity provides remarkable robustness. For example, after a stroke, the neural network reorganizes in a new architecture to preserve motor function \cite{ward2005neural}. It can also adapt to sensory deprivation to extract the most information from the remaining senses. This is the case in blind individuals: the processing of auditive information can be partially deferred to the visual cortex, improving their sound localization abilities \cite{gougoux2005functional}.

Neural plasticity can be considered as a part of the whole mechanism governing homeostasis of both shape and function. This conceptual proximity is also justified by biological evidence such as the discovery of the role of electrical activity during morphogenesis. In particular, the same ion channels are used both for local communication between non-neuronal cells during embryogenesis and in the neurons to carry action potentials \cite{levin2003bioelectromagnetics}. More generally, there seems to exist a continuum between the phenomena we usually call growing, learning, and computing. Each of these abilities may be considered a different aspect of the same underlying self-organizing system.

Moreover, there is evidence that the DNA does not encode precise details of the resulting neural network. There is an information gap between the size of the DNA and the complexity of the neural network, generally referred to as genomic bottleneck \cite{zador2019critique}. The DNA only specifies the local behavior of cells through the shape of the proteins it encodes. The neural network is then a structure that emerges through these local interactions and yields useful biological processing of the inputs received by the senses \cite{banzhaf2004challenge}.

Despite the crucial role of growth in the emergence of intelligence, modern advances in artificial neural networks mainly focus on the handcrafted design of static maps of neural connections. During the phase called learning—that is in fact quite far from the biological sense of this word \cite{zador2019critique} —the connections of this architecture are optimized to reduce the error on the task to solve.

Some effort has been made to include an automatic process to incrementally design neural network architectures. These techniques include the use of genetic algorithms \cite{stanley2002efficient}\cite{nadizar2021effect} or the introduction of a growing phase in artificial neural networks \cite{miller2020evolving} \cite{mixter2020growing}. Nonetheless, in these works, the process is a tool for navigating the topological parameter space, rather than treating the learning as a developmental problem.

The developmental problem has also been addressed as an independent task. In this case, the goal is to model the phenomena of morphogenesis observed in living organisms. Successful approaches used cellular automata along with artificial neural networks implementing local rules. They were able to produce complex patterns from localized interactions. \cite{nichele2017neat} \cite{mordvintsev2020growing}.

More radical approaches tried to develop an artificial substrate where life-like phenomena could emerge in an open-ended environment \cite{gregor2021self} \cite{chan2020lenia}. Despite their great promise, we are still far from recreating the whole evolution process that gives rise to intelligence.

In this work, we attempt at bridging the gap between growth and computation by using neural cellular automata (neural CA) \cite{mordvintsev2020growing}. Neural CA is a spatially distributed system composed of cells that interact through local interaction. Their update rule consists of an artificial neural network and thus can be optimized through classical and efficient gradient descent-based techniques. Even if the learning process is still happening through a procedure exterior to the system, the local rules encode both the developmental process—the transformation from a random grid to a configuration suitable for computation—and the information processing itself. We found that the cells were able to transmit and combine in a meaningful way the information from input cells to output cells used as a readout. The system demonstrates long-term stability and robustness to noise and damage. We illustrate these abilities on a simple control task as a proof of concept. 

\section{The pole balancing task}

The cart-pole problem is a commonly used toy problem in the reinforcement learning community. In this environment, an agent controls a cart-pole system. It can observe the pole angle and its angular velocity, the cart position, and its linear velocity. Based on these observations, the agent must decide whether to apply a force on the left or the right of the cart in order to maximize reward, i.e. the time spent with the pole up and the cart in the center. The simulation ends if the cart hits a wall or if the pole falls. We chose this problem because of its low number of inputs and outputs that allow the use of a small-sized grid and an easy experimentation environment.

\begin{figure}
    \centering
    \includegraphics[width=0.5\textwidth]{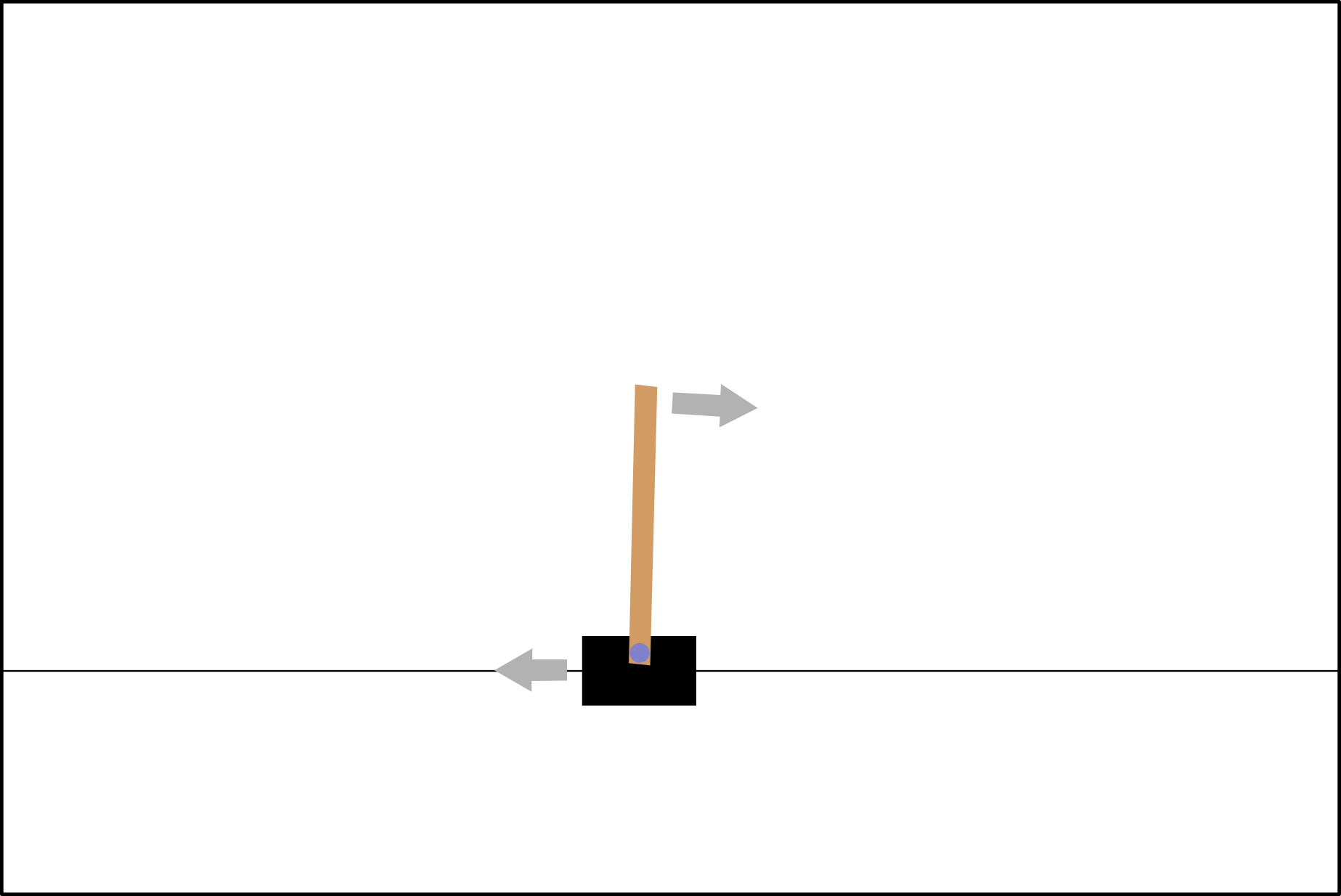}
    \caption{The cart-pole environment. The top arrow represents the angular velocity of the pole, the bottom arrow, the velocity of the cart.  }
    \label{fig:cartpole_envi}
\end{figure}

We used the implementation of this environment provided by the OpenAI gym library\footnote{\href{https://gym.openai.com}{\color{blue}https://gym.openai.com/}}. We modified the reward function to favor agents that stay in the center.  The reward given at time step $t$ is given by the equation (\ref{reward_formula}) where x is the position of the cart, L the length of the track. In the center (x=0), the agent receives 1, the maximum reward, while if it is on the edges of the track (x= ± L), the agent receives 0, the minimum reward. 

\begin{align}
r_t  = \left\{
    \begin{array}{ll}
        \cos\left(\frac{x \pi}{2L}\right)^2 & \mbox{if $t$ is not a final step} \\
        -100 & \mbox{else.}
    \end{array}
\right. \label{reward_formula}
\end{align}

\section{Cell state}

 There are 3 types of cells in a grid: the intermediate, the input, and output cells.

The state of each cell is composed of 6 channels. The first is the \emph{information channel} where meaningful input and output information transit. The third is identifying the inputs: it is equal to 1 in the input cells, 0 elsewhere. The fourth similarly identifies the outputs. The remaining three are hidden channels.

The state of the input cells cannot be changed, the information channel transmits the observation from the environment, and the other channels except the output identifier are set to 1. The values of the information channel of the output cells are used as the output of the system to be optimized to solve the task.

\begin{figure}
    \centering
    \includegraphics[width=0.7\textwidth]{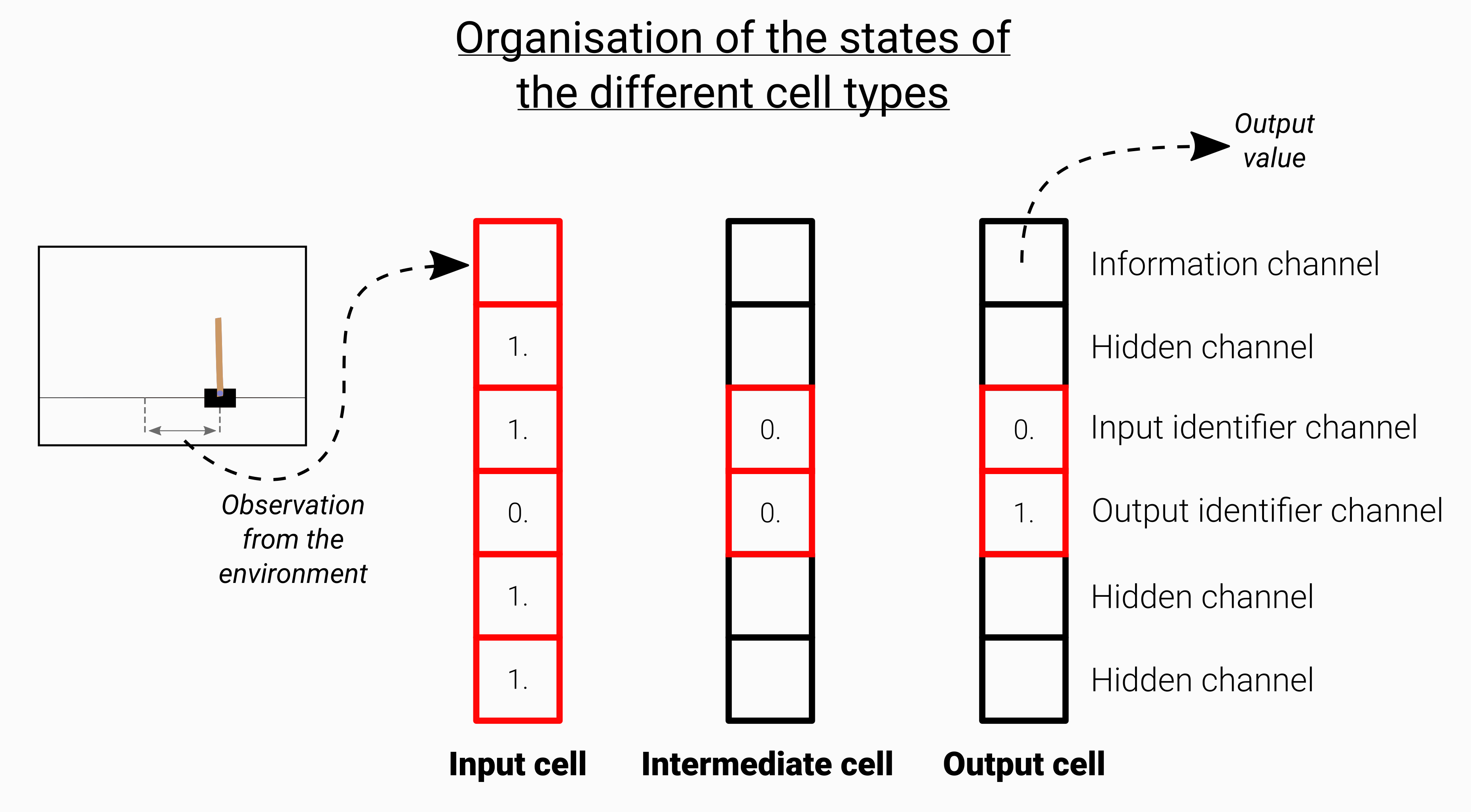}
    \caption{The role of the different channels and types of cells. Channels that are fixed are represented in red and in black are channels that can be modified by the update rule. }
    \label{fig:cell_states}
\end{figure}

The meaning of each channel and the different types of cells are represented in the figure \ref{fig:cell_states}.

\subsection{Input encoding}

 We use redundancy in the inputs: each of the 4 physical observations of the environment is linked to 2 input cells. We thought it could improve the opportunity for information combination and robustness. Note that the type of information contained in the input is not directly provided. To know which observation each input cell encodes, the CA must rely on the spatial position of the inputs or the value of the information channel.

The value of each observation is scaled by a constant factor before being transmitted to the input cell. The choice of the factor corresponding to each observation was chosen to get similar ranges of values in the information channel. 

\subsection{Cell position}

\begin{figure}
    \centering
    \includegraphics[width=0.5\textwidth]{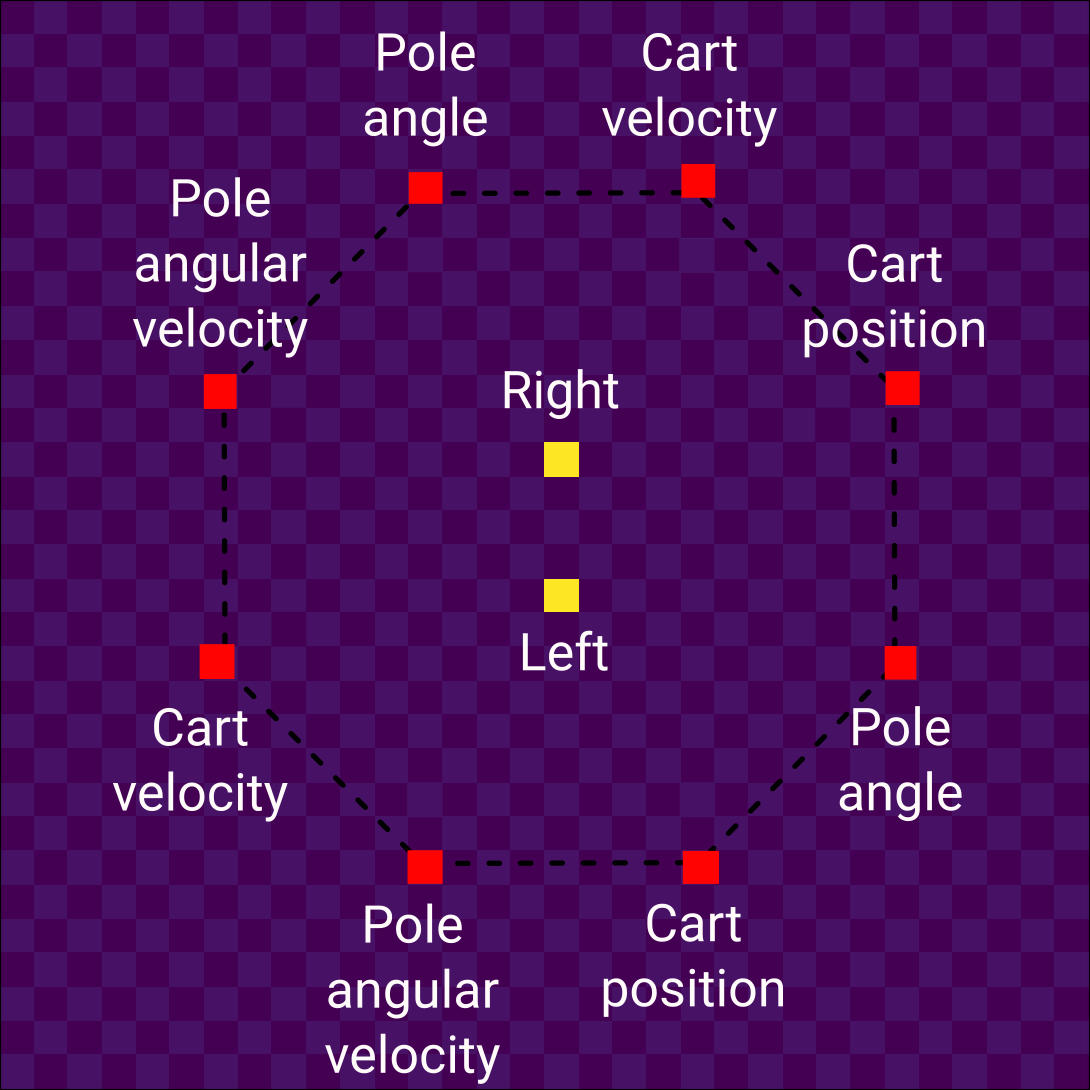}
    \caption{The position of input (in red) and output cells (in yellow).}
    \label{fig:io_positions}
\end{figure}

The 8 inputs are arranged in an octagonal shape (dotted line) on a 32x32 grid with zeros at the boundaries as shown in figure \ref{fig:io_positions}. The two output cells are offset by 2 cells from the center of the octagon. We chose this configuration to ensure an almost equal distribution of the distance between each input and output. 

\section{Model}

 Except for the design of the cell states, the neural CA architecture we used is similar to the one described for the self-classifying MNIST task \cite{randazzo2020self}. The perception layer is composed of 20 learnable 3x3x6 filters, and the single hidden layer counts 30 units. In total, our model has 1854 learnable parameters.

\begin{figure}
    \centering
    \includegraphics[width=0.9\textwidth]{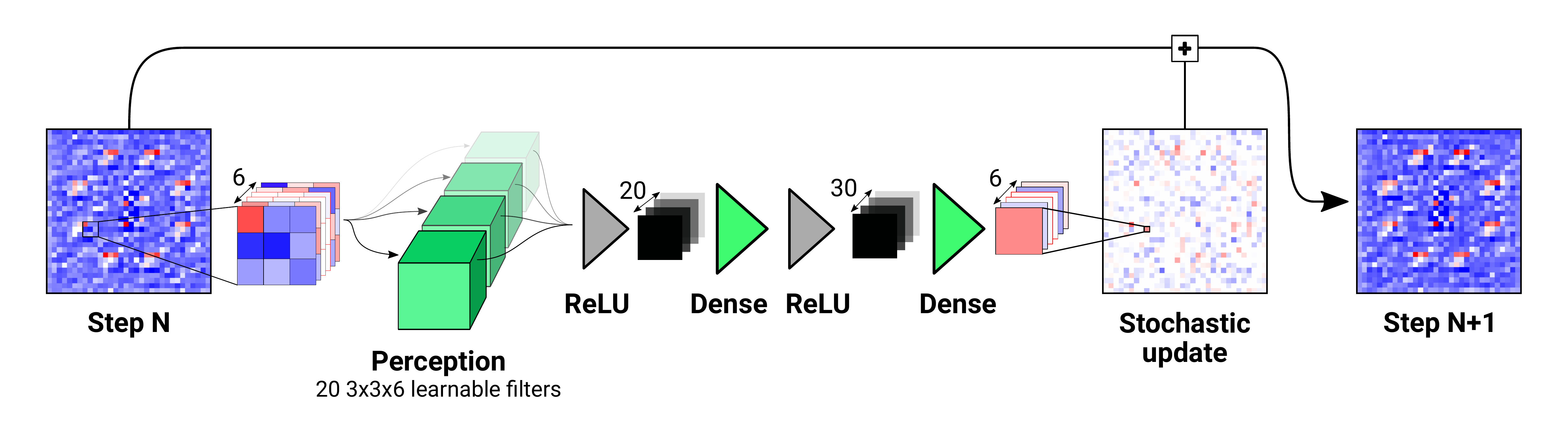}
    \caption{The architecture of our model. Positive values are depicted in red, negative in blue. Green objects identify the learnable parameters of the model.}
    \label{fig:model_archi}
\end{figure}

As in the original model, the update rule is stochastic: at each step, each cell has a 0.5 probability of being updated. This choice is made to avoid temporal synchronization that relies on a centralized clock. The figure \ref{fig:model_archi} summarizes the architecture of our model. 

\section{Training procedure}

Our model can be abstracted as a black-box function that takes inputs (that will be fed to the information channel of input cells) and transforms them into outputs (the information channel of output cells). This function is differentiable with respect to its parameters (the neural network used as the update rule) and thus can be optimized with classical gradient descent techniques. In this case, we used the Adam optimizer \cite{kingma2014adam} provided by the TensorFlow library\footnote{\href{https://www.tensorflow.org/}{\color{blue}https://www.tensorflow.org/}}. 

\subsection{Algorithm}

To tackle the cart-pole problem, we used Deep Q-learning \cite{mnih2013playing} where the usual artificial neural network is seamlessly replaced by a neural CA. The deep Q-learning algorithm aims at approaching the expected reward given a state and an action. More precisely, the function to be learned is given by equation (\ref{q_value_eq}) where $t$ is the index of the current time-step. $s_t$ is the current state and at the action to evaluate. $r_t$ is the reward and $\gamma $ is a discount factor. 

\begin{align}
    Q(s_t, a_t) &= E[r_{t+1} + \gamma r_{t+2} +\gamma^2 r_{t+3}+ ...|s_t, a_t ] \label{q_value_eq}
\end{align} 

To keep the CA values in the information channel in a range coherent with the other cells, we scale the outputs of the CA by a factor of 100 to get the Q-value estimates. 

\subsection{Loss function}

The loss function for the task is the L2 loss between the output and the target. To achieve long-term stability, we added a penalty for cells that have channel values out of bound [-5,5]. Note that excepted this overflow condition, the states of the intermediate cells are not directly optimized, they are free to evolve insofar as their influence on the outputs reduces the error. The formula used to compute the loss is given in equation (\ref{loss_eq}) where $N$ is the size of the grid, $\lambda$ a parameter to control the amount of overflow penalty.

\begin{align}
    \textrm{Loss} &= \textrm{L2}(outputs, target) + \frac{\lambda}{N^2} \sum_{i,j = 1}^{N}{\sum_{chan=1}^{6}{ (\textrm{clip}(Grid_{i,j,chan}, -5,5) - Grid_{i,j,chan})^2  }}  \label{loss_eq}
\end{align} 

\subsection{Robustness}

\subsubsection{Damage}

\begin{figure}
    \centering
    \includegraphics[width=0.5\textwidth]{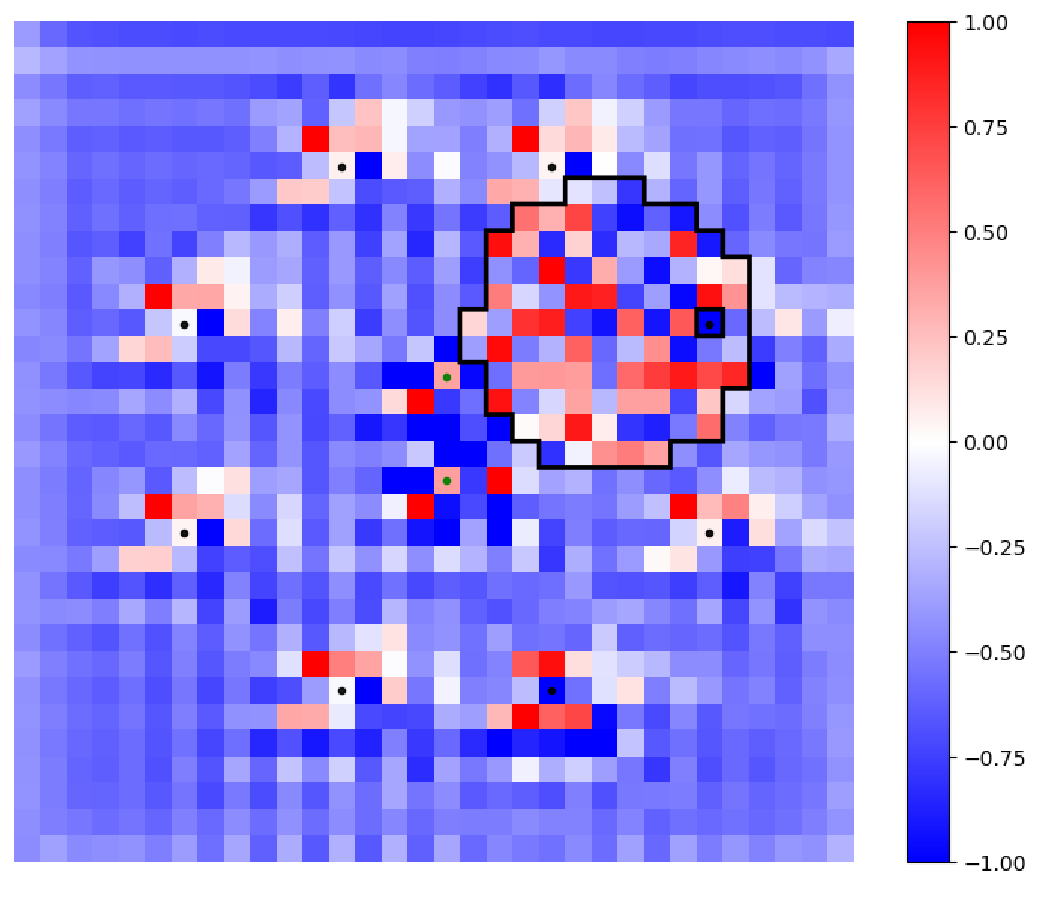}
    \caption{A damaged grid shown on the information channel. The shape of the damaged region is an irregular circle due to rasterization effects.  }
    \label{fig:damage}
\end{figure}

To increase the robustness of the system, we damage half of the grids present in the pool. Damage consists of a circle of the grid replaced by uniform random values in [-1,1], as shown in black in figure \ref{fig:damage}. Note that damage impacts all the channels that can be modified and that inputs are not affected by damage while outputs are. 

\subsubsection{Noise}

Before applying each update, we perturbed it with uniform noise. Following what was done in \cite{randazzo2020self}, we used a noisy update to favor a long-term stabilization of the cell states.

\subsection{Neural CA training}

 As introduced in \cite{mordvintsev2020growing} we used pool sampling for the states of the neural CA during training to learn persistent behavior.

As described in the deep-Q learning algorithm, we alternate phases where we explore the environment by letting the neural CA controls the cart-pole agent and by taking random actions; and training of the neural CA using the target values based on the rewards stored in the memory of the agent. 

\subsubsection{Environment exploration}

 To get a stable long-term behavior of the cart-pole agent we did not use a fixed horizon for the environment. Instead, we use pool sampling also for the states of the cart-pole, as done for the neural CA grids.

The probability of taking a random action is given by the parameter $\epsilon$ that is decreased during the training of the agent, as in the original deep-Q learning algorithm. The exploration of the environment begins by sampling a grid from the grid pool, a cart-pole state from the pool of environment states. Then we let the neural CA, starting from the sampled grid, evolve for a random number of steps from 50 to 60. After, we choose the action that corresponds to the greatest output of the neural CA and obtain a new environment state. We put the grid back in the grid pool and sample a new one. We repeat this operation for K environment steps.

\begin{figure}
    \centering
    \includegraphics[width=\textwidth]{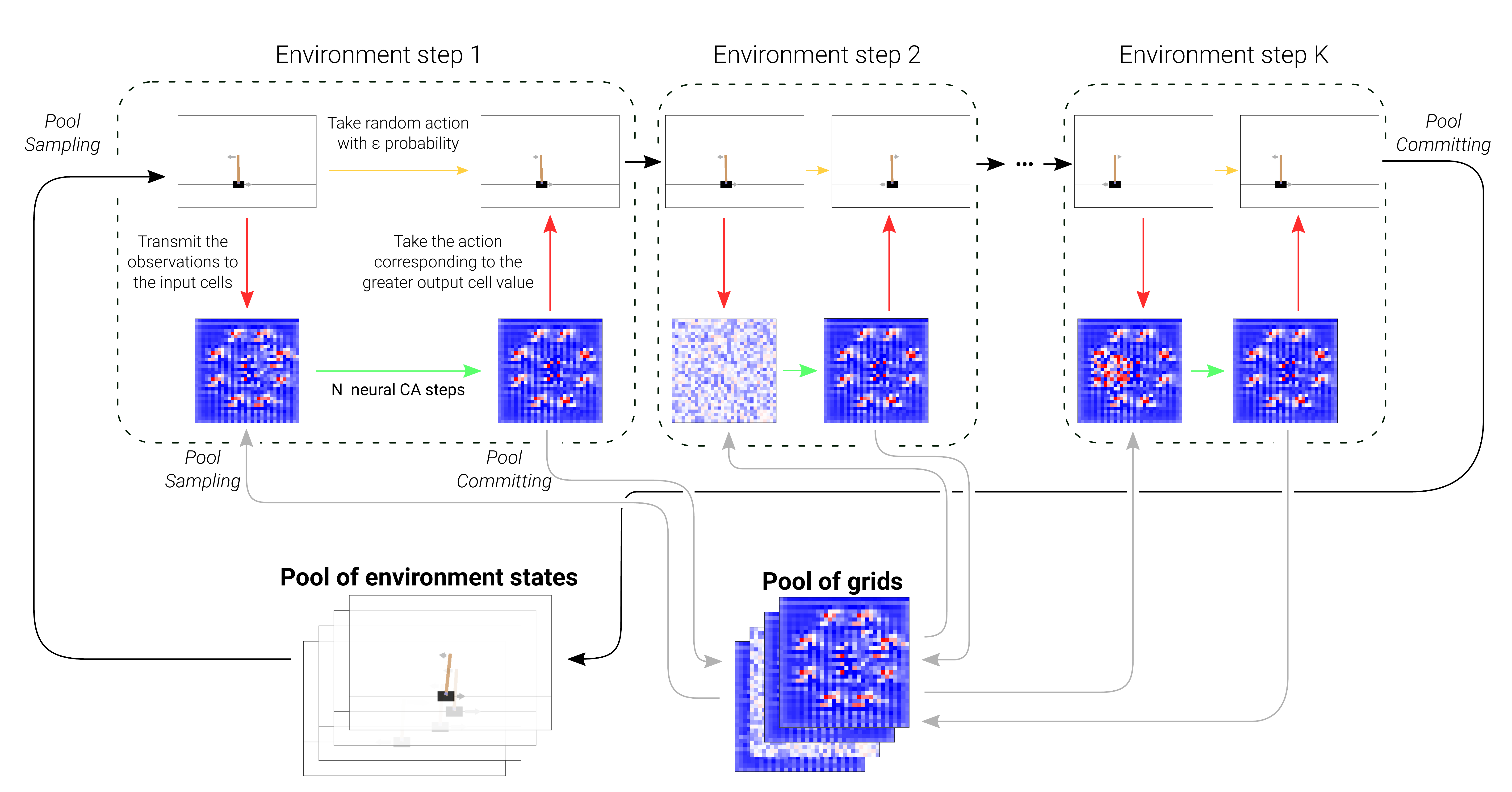}
    \caption{The procedure for the exploration phase. We match cart-pole states and grids randomly sampled from two independent pools. This provides a way to simulate long-term dynamics both for the cart-pole environment and for the neural CA. In practice, we used K=2. }
    \label{fig:training_proc}
\end{figure}

If the environment ends, we reset the environment to reach the end of the K steps. After the K steps, the state of the cart-pole is committed in the pool of environment states. We also randomly replace grids by the initial state to be sure that the neural CA always keeps the knowledge of how to start from a raw grid. The procedure is illustrated in the figure \ref{fig:training_proc}. 

\subsubsection{Training}

Between each exploration phase, we sample several batches of transitions $(o_t, a_t, r_t, o_{t+1})$—where $o_t$ is the observation at time $t$, $a_t$ the action taken, and $r_t$ the reward received—from the memory of the agent that were stored during the exploration phases. We train the neural CA according to the expression of the target value and the error to optimize given by the deep-Q learning algorithm (equation (\ref{loss_for})). Then, each transition is matched with a neural CA grid randomly sampled among the pool of grids that we let evolve for 50 to 60 steps. We next compute the loss on the final state of the grid, according to the formula (\ref{loss_for}) where $y_i$ is the target value for a given transition at time $j$ sampled from the memory of the agent, and $Q(o_j, a_j;\theta)$ is the output of the neural CA for an action $a_j$ and an observation $o_j$. $\theta$ are the parameters of the update rule to be optimized.  
We perform a gradient step on a batch composed of 16 such grids. As described in \cite{mordvintsev2020growing}, back-propagation through time is used to compute the gradient of the loss with respect to the parameters of the update rule. 

\begin{align}
&\bullet ~ y_j  = \left\{
    \begin{array}{ll}
        r_j & \mbox{if $j$ is a final step} \\
        r_j + \gamma* \textrm{max}_{a'} Q(o_{j+1},a';\theta) & \mbox{else.}
    \end{array}
\right. \\
&\bullet ~ \mbox{$TaskLoss$ = $(y_j - Q(o_j, a_j;\theta))^2$} \label{loss_for}
\end{align}

The training procedure runs for around 15k gradient descent steps and 3k environment steps. The training took between 20 min and 1H on a GPU. We used a learning rate of 5e-3 that decays to 5e-4 and then to 5e-5 after respectively 1000 and 10000 steps. Note that the hyperparameters used the training were not optimized and we mainly aimed at solving the task, not efficiency. 

\subsection{Model initialization}

 We found that when trained directly for the task, the model was trapped in a local minimum where it outputted constant values, no matter the state of the inputs. We think that this is because there need to be iterated applications of the update rule on each of the intermediate cells between the inputs and outputs to transmit and modify the information. This repeated use of a neural network makes the gradient vanish, as observed in vanilla Recurrent Neural Networks \cite{hochreiter1998vanishing}.

To solve this problem, we first trained the neural CA on an easier task: both outputs were optimized to compute the mean of the inputs. We found that it was able to learn with a reasonably low error after several thousand gradient descent steps.

This initialization enables the neural CA to learn to stabilize the states of the cells, make an information link from input to output, and a linear combination of the input values at the output cells. This procedure is similar to what is used in curriculum learning \cite{bengio2009curriculum} where an easy subset of the task is learned before tackling the whole problem. Here we did not use a sub-task as a starting point but a different task that shared common requirements.

The whole training procedure can be reproduced online in a \href{https://colab.research.google.com/github/aVariengien/self-organized-control/blob/main/code/Towards-self-organized-control-notebook.ipynb}{\color{blue}Google Colab notebook}.

\section{Results}

\subsection{From estimating Q-values to stable behavior }

To get a persistent control of the cart-pole agent, we begin by transmitting the observation of the current state in the input cells. We let the neural CA evolve for a random number of steps between 50 and 60. We take the action corresponding to the maximum output value and we input the new observation to the neural CA. The grid is initialized with uniform random values and the same grid is used during the whole simulation. After training, the cart-pole controller with neural CA is tested for how long the pole can remain balanced. Moreover, we verify its resistance to damage, noise, and input deletion. 

Our model was able to solve the cart-pole problem and achieve long-term stability of both the pole balancing and of the states of the CA. It was able to balance the pole for more than 10k simulation steps. Detailed performance evaluation can be found in table \ref{perf_table}.

\subsection{Resulting neural CA }

Beyond performance, it is interesting to visualize the activities of the neural CA during the control of the cart-pole agent.

\begin{figure}
    \centering
    \includegraphics[width=\textwidth]{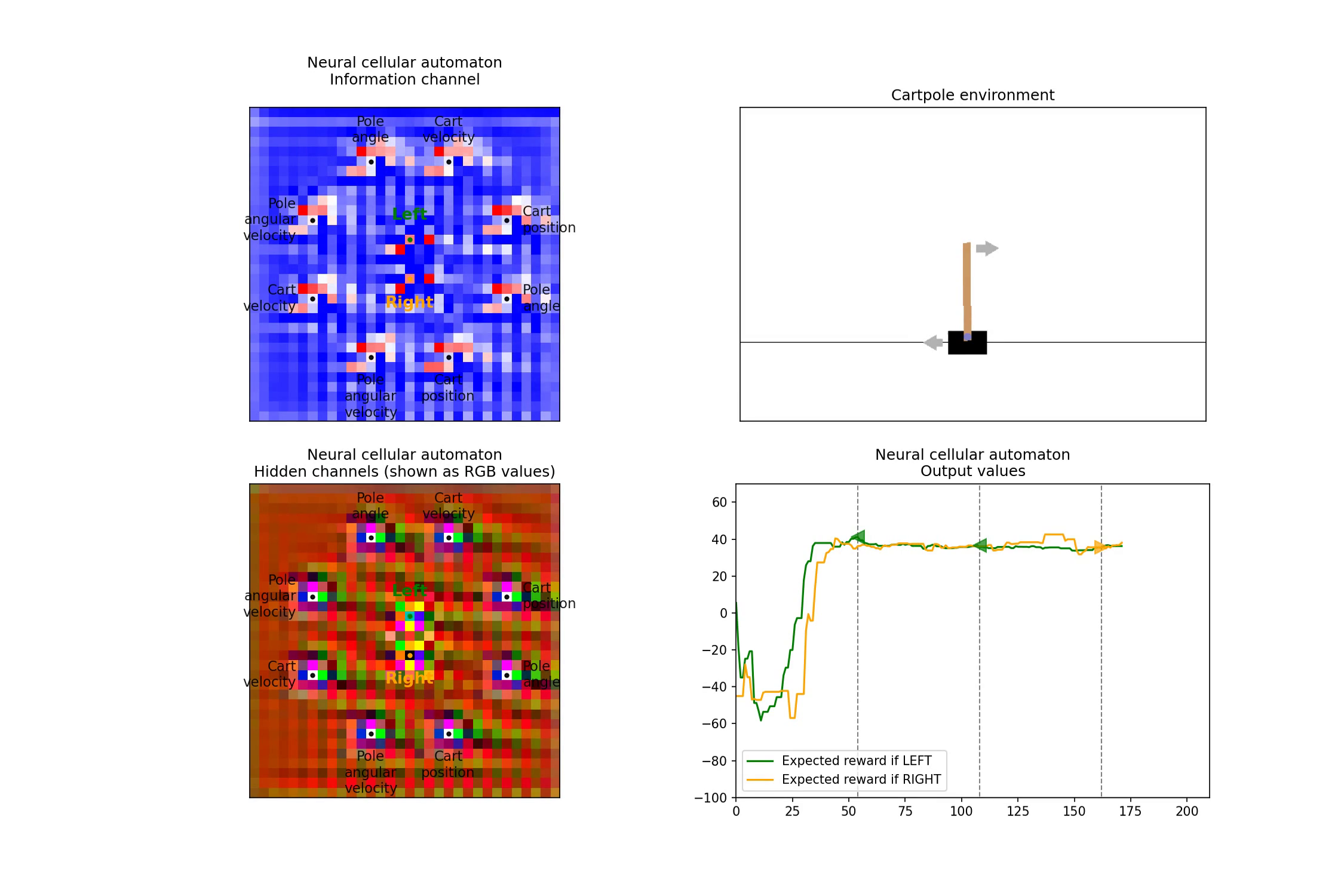}
    \caption{On the left, the states of the neural CA. For the information channel, negative values are in blue, positive in red while the hidden channel as represented as RGB values. Bottom right: the plot of the output values of the neural CA. The vertical dotted lines denote when the action that has the maximum expected reward is taken by the cart-pole agent. Green triangle: "push left" action, orange triangle: "push right" action.  Videos of the cart-pole agent and the neural CA in action are available at \href{https://avariengien.github.io/self-organized-control/}{\color{blue}https://avariengien.github.io/self-organized-control/} }.
    \label{fig:nca_envi_no_noise_no_damage}
\end{figure}
 
 In figure \ref{fig:nca_envi_no_noise_no_damage}, we observe that the first 50 steps lead to a precise spatial organization of the grid and to the stabilization of the output values. Once stabilized, this global shape will not change during the whole run. This can be thought as the developmental part of the neural CA. This phase can be seen in the videos on the interactive version of this preprint available at \href{https://avariengien.github.io/self-organized-control/}{\color{blue}https://avariengien.github.io/self-organized-control/}.
 
\begin{figure}
    \centering
    \includegraphics[width=\textwidth]{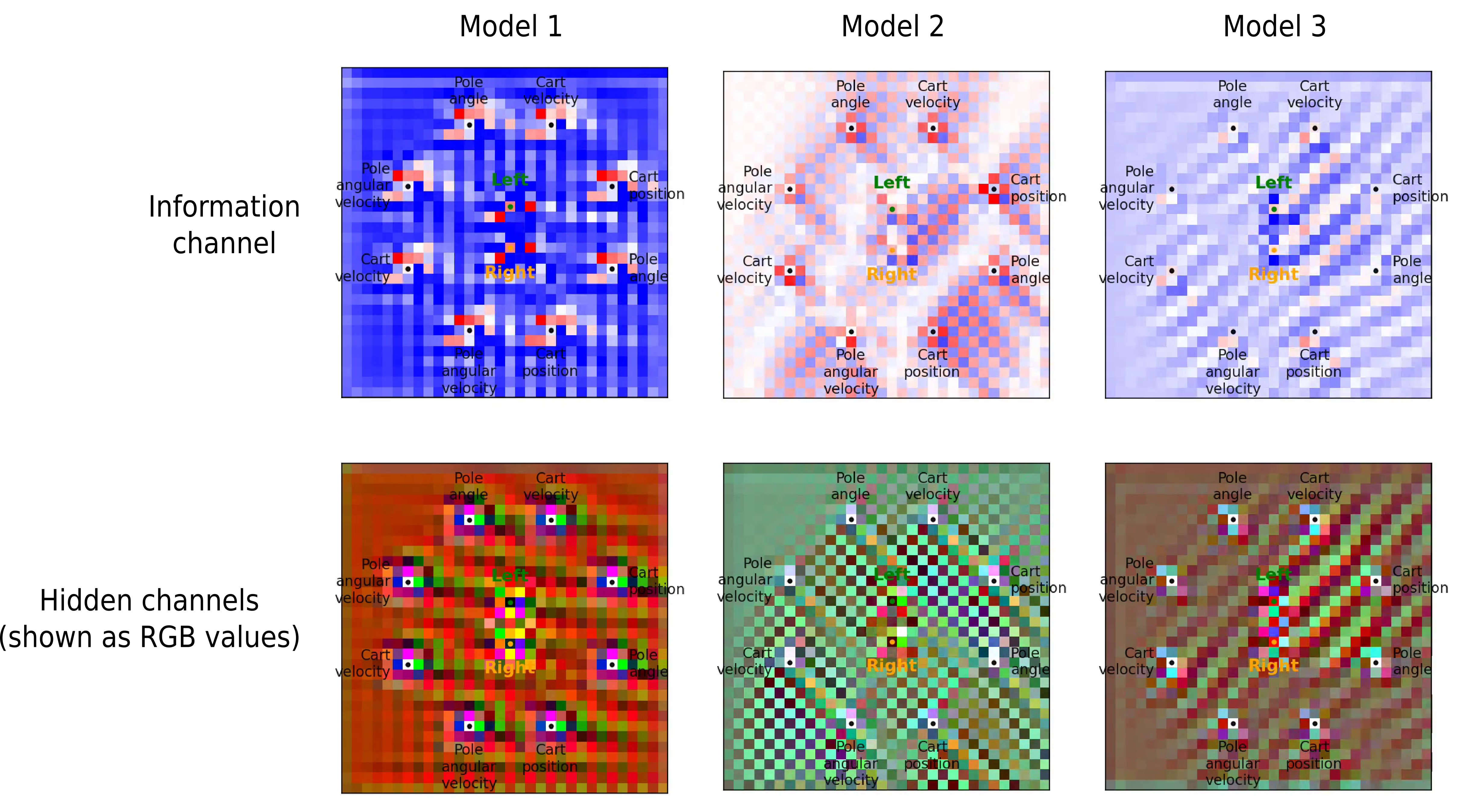}
    \caption{The stable spatial organization of the grid for 3 independently trained neural CA. Their dynamics can be observed in videos available at \href{https://avariengien.github.io/self-organized-control/}{\color{blue}https://avariengien.github.io/self-organized-control/} }.
    \label{fig:diversity}
\end{figure}

Then, during the remaining part of the simulation, the spatial activity is changing in phase with the physical observations. This is the computing phase.  Even if the two phases seem to exist in the different models we found, the exact organization of the grid differs significantly, as visible in figure \ref{fig:diversity}. It is exciting to see that a wide variety of shapes emerges from optimizing for the same function! 

Note that the output values are always really close to one another. Since the pole is in a balanced state, the difference between the expected reward after going left or right is small. Going left then right or going right then left will not yield a great difference in total reward. 

\subsection{Robustness abilities }

During training, the neural CA has always at least 50 steps between the update of the inputs, where damage can occur, and the readout of the output values. During testing, we also experimented with a more challenging type of damage we called \emph{uniformly distributed damage} where at each CA step the grid has a constant probability of being damaged. 

 Because this type of damage was more difficult to cope with, we decreased the damage frequency: on average, the CA receives one damage every 4 input updates with uniformly distributed damage and one every 2 input updates with the damage used during training.

The performance of the neural CA with different perturbations is summarized in the table \ref{perf_table}. The score denotes the number of environment steps before the pole falls, or the cart hits a wall. In each situation, we computed the mean score on 100 independent runs, as well as the standard deviation. To ease the analysis, we conducted the experiments of this section only on a single model, nonetheless the main conclusions generalize to the other models. 

\begin{table}
 
  \centering
  \begin{tabular}{l ccc}
    \toprule

    ~ & No damage & Damage after input update & Uniformly distributed damage \\
    \midrule
    No noisy update    & 13273 ± 11905 &  2598.19 ± 2241  & 391.6 ± 283 \\
    With noisy update  & 1296.3 ± 899 &  739.7 ± 473  & 345.7 ± 214  \\
    \bottomrule
  \end{tabular}
  \vspace{2mm}
  \caption{Performance of the cart-pole agent with several types of perturbation. The score is the number of time steps the cart-pole stays balanced without hitting walls. The scores are averaged on 100 runs and are noted $mean$ ± $standard$  $deviation$. }
  \label{perf_table}
\end{table}

\subsection{Resistance to damage }

\begin{figure}
    \centering
    \includegraphics[width=\textwidth]{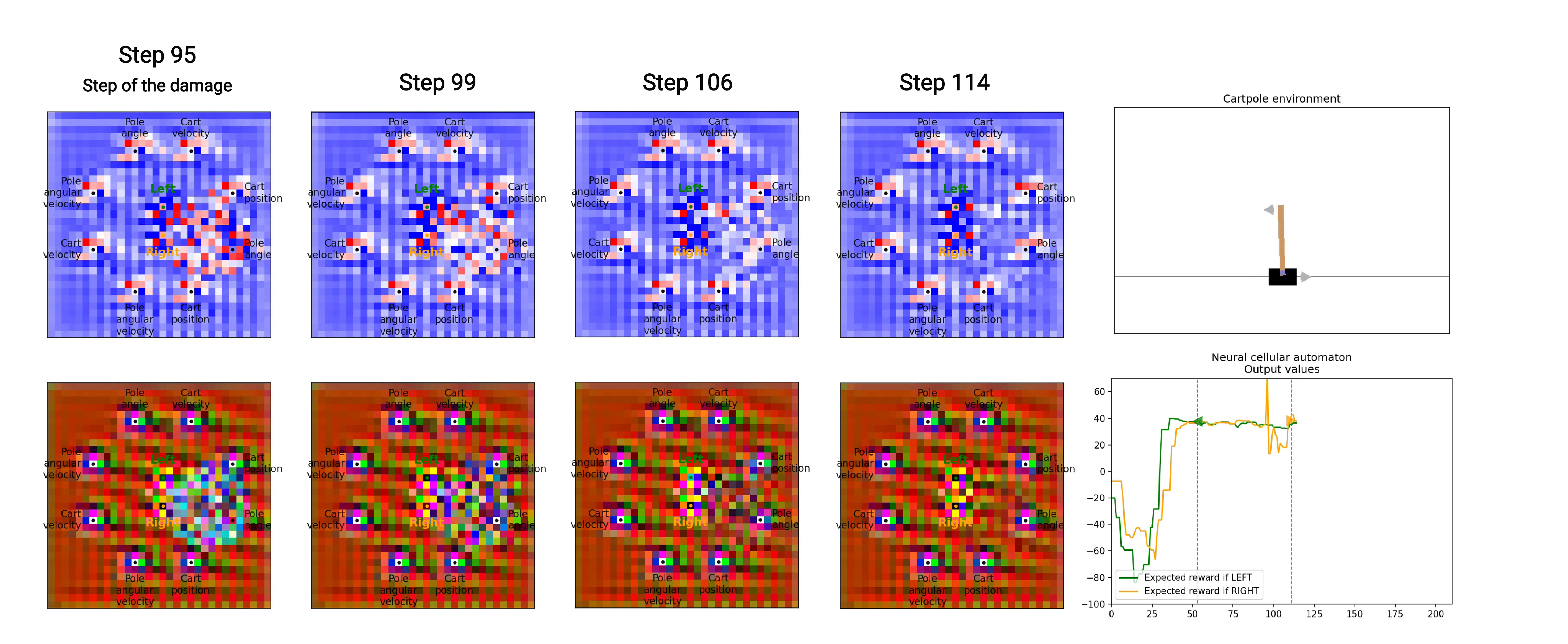}
    \caption{A neural CA controls a cart-pole agent and recovers from grid damages. The evolution of the grid after damage is depicted. Videos of the recovery can be found at \href{https://avariengien.github.io/self-organized-control/}{ \color{blue}https://avariengien.github.io/self-organized-control/}}.
    \label{fig:damage_recovery}
\end{figure}

 We found that the neural CA was able to maintain its shape and its function despite frequent damage. In the figure \ref{fig:damage_recovery} we can observe how the grid recovers its shape after damage. Although damage can lead to great perturbations in the output values and so to random actions, the agent is still able to stabilize the pole for several hundred steps. 

Moreover, the neural CA was not trained to recover from uniformly distributed damage, this explains the greater diminution in the average score visible in the table \ref{perf_table}.

\subsection{Resistance to noise  }

\begin{figure}
    \centering
    \includegraphics[width=\textwidth]{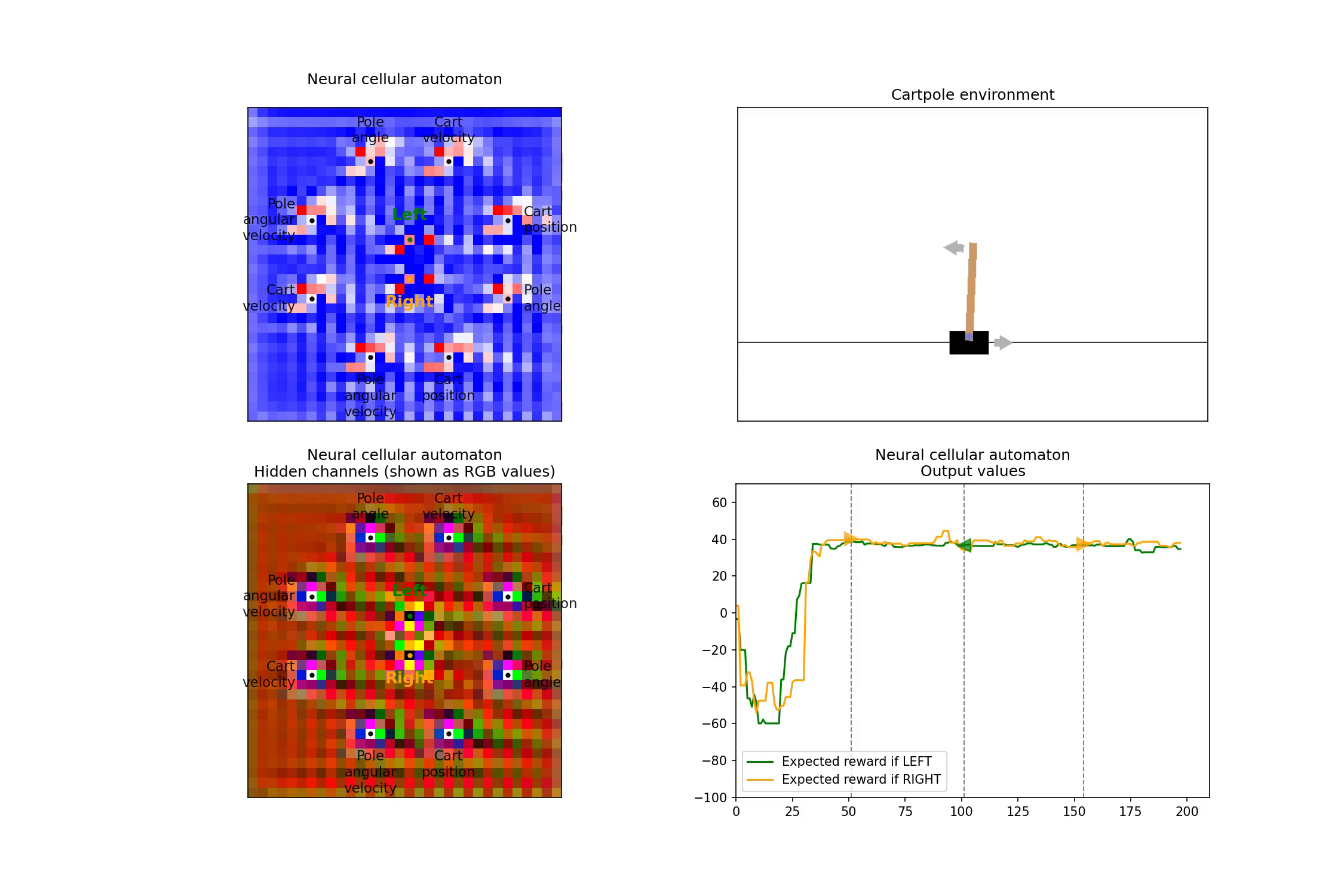}
    \caption{A neural CA controls a cart-pole agent  with noisy update.  }
    \label{fig:nca_envi_noise_no_damage}
\end{figure}

The amount of noise added to each update is often of the same order as the difference between the two outputs when the pole is in a balanced state. This is why we observe in the figure \ref{fig:nca_envi_noise_no_damage} the green and orange curves subject to stochastic variations that lead them to cross many times between each readout. The policy that controls the cart-pole agent is thus heavily randomized. Despite the noisy update, the neural CA can produce a probability distribution of actions such that a stable behavior emerges. 

\subsection{Resistance to input deletion }

One of the particularities of this neural CA model is its flexibility. For instance, the number of inputs and outputs can vary without changing the architecture. We only have to replace input or output cells with intermediate ones.

We were interested in exploring this flexibility and whether our model showed robustness to input deletion. Because observations from the cart-pole environment are encoded redundantly, we tested if it was able to exploit this particularity even if it was not trained for this. 

In the table \ref{table_input_deletd}, we can observe the consequences of deleting each input. 
We computed the mean scores on 25 independent runs for each input deletion. 
Each column corresponds to one observation type, each row corresponds to the first or the second input cell encoding this observation being deleted. 
\begin{table}
 
  \centering
  \begin{tabular}{l cccc}
    \toprule

    ~ & Cart position & Cart velocity & Pole angle & Pole angular velocity \\
    \midrule
    First input deleted    & 814.1 ± 632 &  293.4 ± 144  & 924.6 ± 601 & 267.2 ± 139 \\
    Second input deleted  & 814.6 ± 608 &  168.2 ± 46  & 53.6 ± 37 & 103.2 ± 30  \\
    \bottomrule
  \end{tabular}
  \vspace{2mm}
  \caption{Mean score and standard deviation of 25 independent runs after each input deletion. The CA were perturbed with damage after the update (on average one every 2 cart-pole steps) and noisy update. }
  \label{table_input_deletd}
\end{table}

 The system seems to be dependent on a few input cells that seriously impair performances such as the second input cell encoding for pole angle and the ones corresponding to the angular velocity, while others seem not to significantly affect its abilities. We hypothesize that even if it has not been directly trained to be robust to input deletion, the robustness to damage and noise includes also adaptation to unseen perturbation.

\begin{figure}
    \centering
    \includegraphics[width=0.8\textwidth]{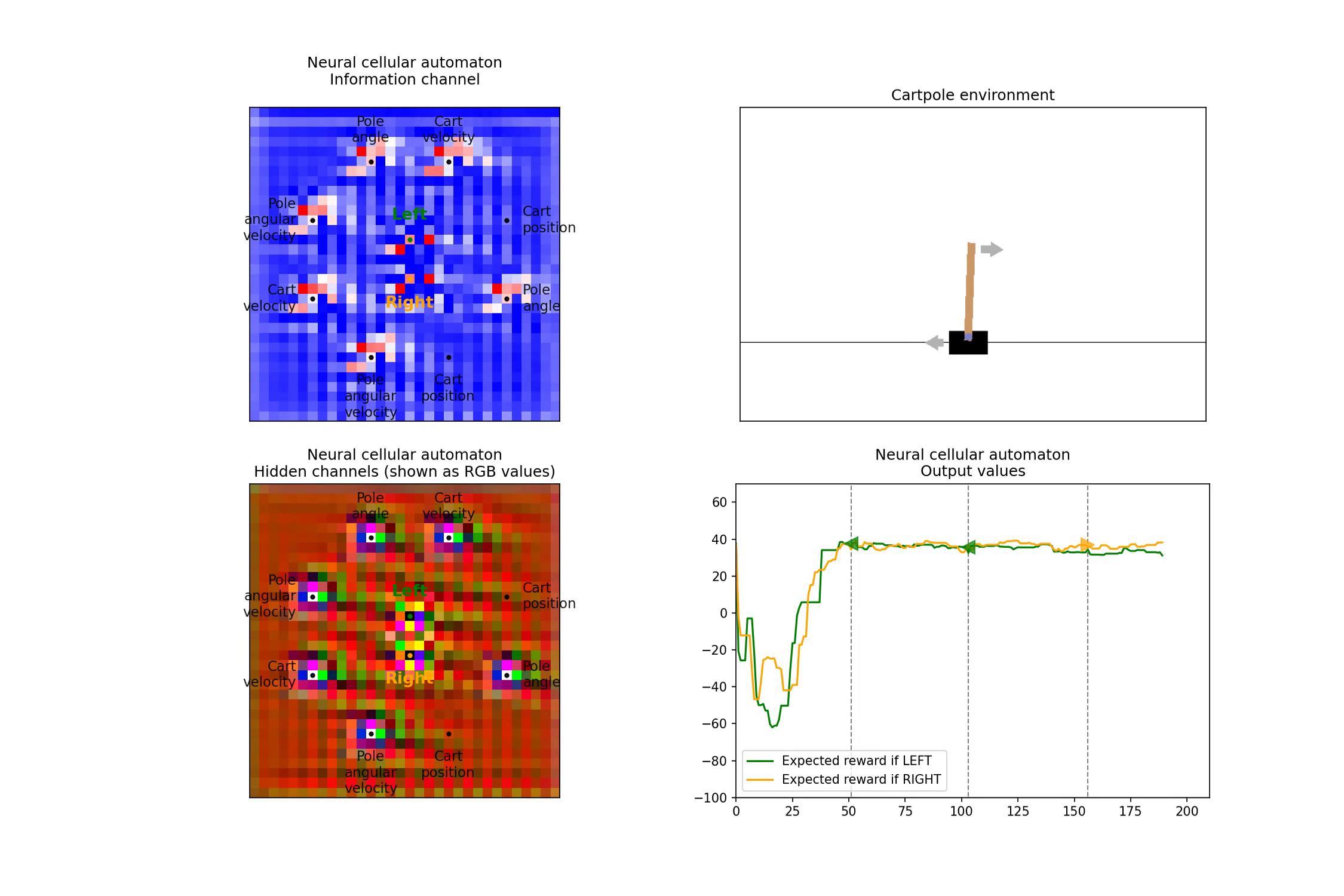}
    \caption{The neural CA and the cart-pole environment it controls. The input cells corresponding to the cart position have been deleted and replaced by intermediate cells. No noise nor damage was added.  }
    \label{fig:nca_envi_inputs_deleted}
\end{figure}

It seems that the inputs corresponding to the cart position do not disturb the control abilities. So we experimented with how the system will react to sensory deprivation by removing these two input cells such that the system has no longer access to this observation. It is still able to maintain the pole balanced for several thousand steps (score of 3926.7 ± 2383 on 25 runs without noise and damage). The reconfiguration of the grid can be observed in the figure \ref{fig:nca_envi_inputs_deleted}. 

\section{Influence field visualization}

In the videos showing neural CA and the environment side to side, we can observe that the regions around input cells are producing a dynamic pattern in phase with the movements of the cart-pole. We develop a visualization tool to investigate the region of neural CA that is influenced by a particular input. To this end, we compared the evolution of the neural CA between a baseline case and a case where a particular input was perturbed. We then computed the relative mean of the difference for each of the cells in the grid, according to the formula (\ref{deviation_formula}). This process is repeated on different observations sampled from the environment and for several grids to get consistent patterns. 

The expression of the deviation used to quantify the influence of a given input on the other cells is given in equation (\ref{deviation_formula}). The norm is the L2 norm and is computed by treating each cell as a 6-dimensional vector. In practice, the mean was computed for 50 different observations, and for each observation, we used 4 independent grids.

\begin{align}
    Deviation = \frac{\textrm{Mean}( \textrm{Norm} (Grid_{baseline} - Grid_{perturbed} ))}{\textrm{Mean}( \textrm{Norm} (Grid_{baseline}),\textrm{Norm} ( Grid_{perturbed} ))} \label{deviation_formula}
\end{align}

\begin{figure}
    \centering
    \includegraphics[width=0.9\textwidth]{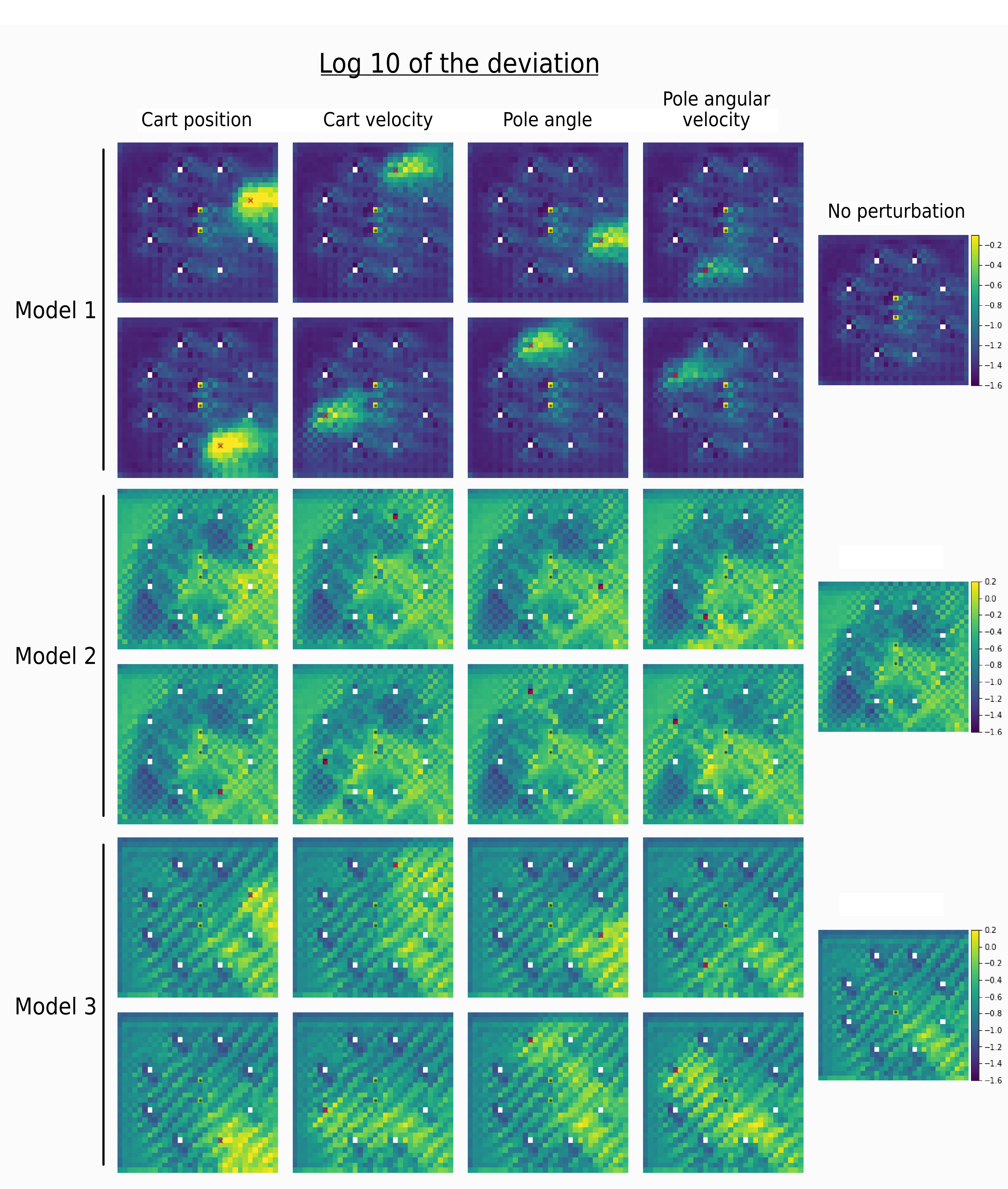}
    \caption{Visualization of the region of influence of each input cell. We plotted the decimal logarithm of the deviation between a standard input and a perturbed one. The perturbed input cell is symbolized by a red cross, black dots identify the output cells. The natural deviation without any perturbation, due to the stochasticity of the system is shown on the right. These results are reproducible in this \href{https://colab.research.google.com/github/aVariengien/self-organized-control/blob/main/code/Towards-self-organized-control-notebook.ipynb}{\color{blue}Google Colab notebook}. }
    \label{fig:influence_field}
\end{figure}

We used as a perturbation the multiplication by a random number between -1 and 1. This ensures that the input will not be out of the range of the possible values, while allowing for a sufficient range to get interpretable visualization. We experimented with different types of perturbation, the resulting visualizations were similar. Each input cell is perturbed independently: its sister input cell transmitting the same observation is not affected by the perturbation. The region of influence for each cell is visualized in the figure \ref{fig:influence_field} for 3 different models. 

 For the model 1, we can observe a localized influence of the input cell with a tendency to be directed toward the right. We also discovered that the inputs that cause the less performance loss if deleted (see table \ref{table_input_deletd}) were the ones positioned on the right. We hypothesize that the input cells on the right side of the grid had less influence on the outputs because the CA learned a rule that can be summarized as "propagate information toward the right". This is allowed by the fact that information is redundant and that the majority of input cells on the right hold the same observation as an input cell on the left side.

For the model 2, we observe a great amount of deviation even without any perturbation. This makes it difficult to interpret the results with perturbation. It seems that the influence of a given input cannot be visible by the fact that the values of another cell are affected, but in the way these values change.

The model 3 could be the intermediate between the two precedents. It presents more deviation without perturbation, while still exhibiting a clear increase in deviation localized around perturbed inputs.

The conclusions that can be drawn from these visualizations are still limited and must be taken carefully. This technique is shared as an attempt to understand the underlying dynamics of the resulting self-organizing system. We think that the development of visualization tools could be a useful step to direct the future design of self-organizing systems. 

\section{Related works}

 The idea that a controller can emerge from a self-organizing system is not new. One of the most studied examples concerns the gait transition in animals i.e. going from walking to running when the velocity of the motion increases. The different limb coordination strategies observed during each gait do not seem to be the result of a control plan transmitted by the brain. Instead, this phenomenon has been described as a phase transition in the self-organizing system composed of the bones, nerves, and muscles used for locomotion \cite{diedrich1995change}. This has notably been modeled by coupled differentiable equations describing mechanical dynamics and neural oscillators to reproduce walking motion \cite{taga1991self}.

More generally, it has been argued that there exists close proximity between goal-directed behavior relying on feedback loops where the agent tries to adjust its action to minimize the distance to its desired state and the dynamics of self-organizing systems. These two models could be different ways of thinking about the same processes \cite{carver2002control}. 

\subsection{Goal-directed cellular automata }

 The artificial design of self-organizing systems has been strongly focused on cellular automata (CA) because of their simplicity and their general abilities. CA have been historically introduced to address general questions about multicellularity in life: how can complex shapes be created from a single cell, maintained, and then replicated?

While the first works focused on handcrafted rules to create self-replicating systems \cite{neumann1966theory}, more recent projects complexified the rules updating the cell states. To search among the wide rule spaces, genetic algorithms have been frequently used to find CA that exhibited a predefined behavior. This enables the design of CA that robustly grows a shape, in effect exhibiting homeostasis \cite{gerlee2017influence}. Another work was able to develop a targeted shape and maintain it despite damage \cite{miller2004evolving}.

Instead of a classical look-up table, some works used update rules implementing more complex algorithms. These types of update rules were used as a generalization of evolvable circuits \cite{bidlo2008instruction} to design CA that performs tasks broader than the historical goal of CA \cite{bidlo2016routine}. As in this case, CA have been used more generally not only for questions related to shape but also for useful decentralized computation \cite{mitchell1993revisiting}.

To improve the search with genetic algorithm and favor the evolvability, some works used variable genotype size \cite{nichele2014evolutionary}. Other works also included developmental function in the CA rules in order to approach the fuzzy, one-to-many, function that maps a genotype and an environment to a phenotype. This was done through the addition of self-modifying abilities in the code of each cell, leading to the creation of self-replicating systems \cite{nichele2016genotype}. 
\subsection{Neural cellular automata }

 Precedent works used evolved neural networks to create CA that grow desired shapes \cite{nichele2017neat}. Then, the introduction of neural CA \cite{mordvintsev2020growing} allowed the optimization of the neural networks used as update rules using the language of differentiable programming instead of genetic algorithms.

This model adds further elements to the questions for which the CA were created. Neural CA enable the creation of self-repairing systems that can grow complex shape from a single cell in 2D \cite{mordvintsev2020growing} or in 3D \cite{sudhakaran2021growing}, and regeneration of functional bodies such as soft robots \cite{horibe2021regenerating}. Beyond investigating homeostasis of shape, neural CA have also been used for decentralized pattern recognition \cite{randazzo2020self} as well as texture synthesis \cite{Mordvintsev2021differentiable}. 

\section{Discussion }
 In this work, we demonstrated that neural CA can be used as a differentiable black-box function theoretically extending its applications to the approximation of any functions. Here we demonstrated its abilities in the context of Deep-Q learning. We used it to solve the simple cart-pole problem. A direct future challenge would be to apply it to more challenging tasks where the input and output dimensionality is much higher.

The computing abilities of the neural CA were maintained over several hundreds of thousand iterations, producing an emergent stable behavior in the environment it controls for thousands of steps. Moreover, the system obtained demonstrated life-like phenomena such as a developmental phase, regeneration after damage, stability despite a noisy environment, and robustness to unseen disruption such as input deletion. In the future, we could also experiment with randomized input and output positions. This would add new challenges: recognize the role of each input and output cell and then create flexible pathways to transmit and combine information.

Even if the developmental phase and the computing phase used the same rules, our system cannot adapt to new environments once the training ends. Future works could explore the possibility of adding plasticity abilities and useful memory of past events stored in the states of the cells. This would mean that the neural CA could recognize a particular situation, and adapt its computations accordingly. Moreover, we envision that even metaplasticity found in biological neurons \cite{abraham2008metaplasticity} could be achieved by neural CA.

Besides the biological plausibility of neural CA, their interest also relies on the fact that they are a highly decentralized computing model. Neural CA could be executed efficiently on dedicated hardware using locally connected microprocessors such as cellular neural networks \cite{cimagalli1993cellular}. Other works explored exciting directions such as framing reaction-diffusion mechanisms as neural CA \cite{Mordvintsev2021differentiable} that would potentially lead to implementation using chemical computing. Those reaction-diffusion systems could also be applied to other tasks than shape homeostasis, such as control.

Beyond the dissociation of the environment and the controller, we could imagine a neural CA that could perform both shape homeostasis and controls the movement of this shape at the same time. If a suitable physical implementation is found, such works could give rise to new robotics and artificial devices with self-organizing abilities that are for now reserved to the living world. 

\section{Additional experiments }

In addition to the cart-pole balancing problem, we explored other tasks and different variations of the neural CA model. Here is a short list of the other tasks we tried that relate to problems solved by biological organisms. To keep this paper short, we chose to focus on a single task, but videos of our additional results and the code to reproduce them can be found at \href{https://github.com/aVariengien/self-organized-control/tree/main/code/AdditionalExperiments}{\color{blue}https://github.com/aVariengien/self-organized-control/tree/main/code/AdditionalExperiments}. 

\begin{itemize}
    \item Exploring an environment to find a target cell: In this task, new cells can grow only next to already living cells. Each cell has an energy value that controls its fire rate. The goal is, starting from a single alive cell, to find a randomly placed target while using in total the lowest amount of energy. 
    \item Following a gradient: This task is similar to the precedent but we provide information for the position of the output. Each cell possesses a read-only channel that is proportional to the distance to the target. This way, the growth can be directed toward the target cell instead of being limited to strategies of random exploration. 
    \item Computing boolean functions: We experimented with computing simple boolean functions such as XOR or its negation, NOT XOR. The environment includes 2 input and 1 output cells. In addition to damage and noise, the position of the input and output cells are randomized such that to solve the task, the cells must communicate without relying on fixed positions.
\end{itemize}

\section{Acknowledgment}

This work was partially funded by the Norwegian Research Council (NFR) through their IKTPLUSS research and innovation action under the project Socrates (grant agreement 270961), and Young Research Talent program under the project DeepCA (grant agreement 286558).

\FloatBarrier
\bibliographystyle{unsrt}  
\bibliography{references}

\end{document}